\documentclass{article} 
\usepackage{iclr2026_conference,times}
\usepackage[most]{tcolorbox}

\usepackage{amsmath,amsfonts,bm}

\def\eqref#1{equation~\ref{#1}}

\def\1{\bm{1}}

\DeclareMathAlphabet{\mathsfit}{\encodingdefault}{\sfdefault}{m}{sl}
\SetMathAlphabet{\mathsfit}{bold}{\encodingdefault}{\sfdefault}{bx}{n}

\usepackage{url}
\usepackage{graphicx, wrapfig, booktabs, tabularx}
\usepackage[colorlinks = true, 
            urlcolor    = teal, 
            citecolor   = brown,
            linkcolor   = teal]{hyperref}
\newcommand\blfootnote[1]{%
  \begingroup
  \renewcommand\thefootnote{}\footnote{#1}%
  \addtocounter{footnote}{-1}%
  \endgroup
}
\title{A circuit for predicting hierarchical structure in-context in Large Language Models}

\author{Tankred Saanum$^{1, *, \dagger}$,  \  Can Demircan$^{2, *}$,  \  Samuel J. Gershman$^{1}$  \& \  Eric Schulz$^{2}$
\\
Harvard University$^{1}$, Institute for Human-Centered AI, Helmholtz Computational Health Center$^{2}$
\\ $^{\dagger}$\texttt{tankredsaanum@fas.harvard.edu}
}

\iclrfinalcopy 
\begin{document}

\maketitle

\begin{abstract}
Large Language Models (LLMs) excel at in-context learning, the ability to use information provided as context to improve prediction of future tokens. Induction heads have been argued to play a crucial role for in-context learning in Transformer Language Models. These attention heads make a token attend to \emph{successors} of past occurrences of the same token in the input. This basic mechanism supports LLMs' ability to copy and predict repeating patterns. However, it is unclear if this same mechanism can support in-context learning of more complex repetitive patterns with hierarchical structure or contextual dependencies. Natural language is teeming with such cases. For instance, the article \texttt{the} in English usually prefaces multiple nouns in a text. When predicting which token succeeds a particular instance of \texttt{the}, we need to integrate further contextual cues from the text to predict the correct noun. If induction heads naively attend to all past instances of successor tokens of \texttt{the} in a context-independent manner, they cannot support this level of contextual information integration. In this study, we design a synthetic in-context learning task, where tokens are repeated with hierarchical dependencies. Here, attending uniformly to all successor tokens is not sufficient to accurately predict future tokens. Evaluating a range of LLMs on these token sequences and natural language analogues, we find adaptive induction heads that support prediction by learning what to attend to in-context. Next, we investigate how induction heads themselves learn in-context. We find evidence that learning is supported by attention heads that uncover a set of latent contexts, determining the different token transition relationships. Overall, we not only show that LLMs have induction heads that learn, but offer a complete mechanistic account of how LLMs learn to predict higher-order repetitive patterns in-context.

\end{abstract}

\blfootnote{$*$ Equal contribution.}
\section{Introduction}\label{section:Intro}

In-context learning is one of the most pervasive features of Large Language Models (LLMs). Informally, in-context learning is simply the ability to predict future tokens more accurately given more contextual information, for instance, feedback, examples, and the like. Crucially, this level of adaptation does not involve a change in model weights but stems solely from the information provided in-context. Consequently, understanding the internal mechanisms that give rise to in-context learning in LLMs, has been a key focus in the machine learning community. A natural starting point of such an analysis for Transformer-based LLMs is the model's attention heads. A conventional generative Transformer Language Model is composed of a sequence of Transformer layers, containing a self attention module, a Multi-Layer Perceptron module, and normalization. The self attention module is responsible for routing information from past tokens to future tokens. In an important study, \citet{olsson2022context} showed that a two-layer (attention-only) Transformer Language Model developed attention heads that made tokens attend to the successor tokens of their past instances. By attending to successor tokens, these attention heads allowed the model to copy previously observed statistical patterns and token bi-grams. These attention heads are called induction heads and have since served as an important explanatory mechanism behind in-context learning.

\begin{figure}[t]
    \begin{center}
    \includegraphics[width=\textwidth]{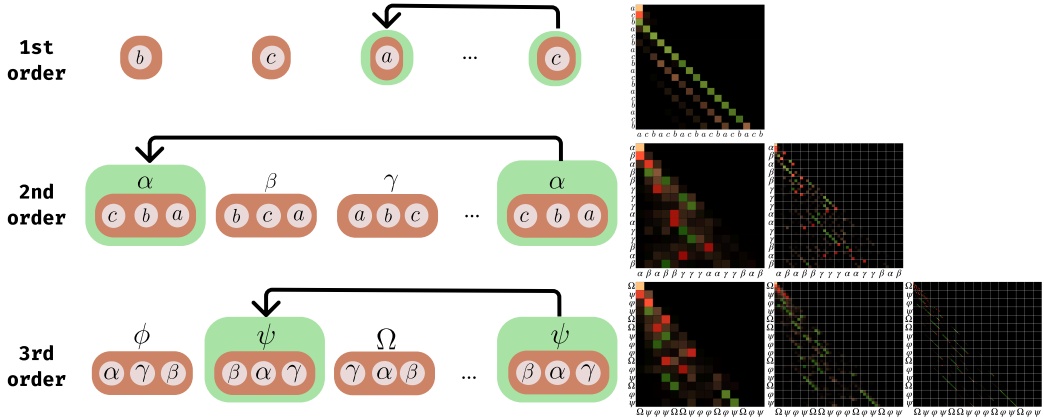}
    \end{center}
    \vspace{-2.5mm}
    \caption{\textbf{Left}: A schematic of our experimental design. First-order structures can be learned by a bi-gram model. Second-order structures introduce context-dependent transition probabilities, requiring the model to identify the current context. Third-order structures generalize this by merging second-order contexts into higher-order ones. \textbf{Right}: Attention patterns from an adaptive induction head. The model moves beyond simple bi-gram behavior by learning to attend to successor tokens in the correct context. Green cells indicate correct attention, while red cells indicate incorrect attention.}
    \label{fig:overview}

\end{figure}

However, induction heads, in their most basic formulation, cannot serve as the be-all and end-all of pattern induction. Many statistical patterns in language and other domains have long-range and hierarchical contextual dependencies. For instance, if a token $x$ has had two different successor tokens in the input, which successor token should it attend to? Natural language is teeming with higher-order repetitive structure where keeping track of bi-gram statistics alone is not sufficient for accurate prediction. For instance, the token \texttt{the} can preface multiple different noun tokens in a text. To predict what token will follow \texttt{the}, we need to consider more than just the set of noun tokens that succeeded it in the past, and integrate higher-order contextual cues. \citet{olsson2022context} indeed reported an induction head that showed signs of context-sensitive adaptation, raising the possibility that induction heads can adapt in-context and actually account for a more substantial part of in-context learning than simple copying. Similarly, \citet{akyürek2024incontextlanguagelearningarchitectures} also report induction heads copying successor tokens based on $2$-gram prefix matching in Transformer Language Models trained from scratch to predict strings from synthetic formal languages.


Expanding on these ideas, we investigate whether induction heads in pre-trained LLMs can learn what successor tokens to attend to in-context when the token sequences have \emph{hierarchical dependencies}. We design a comprehensive set of synthetic token sequences that incorporate repetitive patterns at various levels of hierarchy, along with natural language analogues. In order to predict these token sequences accurately using induction heads, the LLM needs to direct them to attend to specific successor tokens while ignoring others based on these tokens' preceding context. Remarkably, across all LLMs we evaluate, we discover induction heads in later layers that learn what successor tokens to attend to in-context (see Fig. \ref{fig:overview} for task and induction head visualization). We subsequently verify our finding on a simple natural language test, showing how this mechanism is used for sequences closer to the training distribution.


Ultimately, if LLMs learn in-context in virtue of induction heads learning in-context, how do they do it? We propose a simple mechanism explaining how induction heads learn in-context in our task. In our proposed circuit, dedicated heads make tokens attend to the context preceding them, routing information from potentially distant past tokens to allow subsequent context-sensitive attention. From the output of these heads, we can decode whether the current token $x_t$ has the same $N$ preceding context tokens as the \emph{previous} instance of $x_{t'<t}$ after $N$. These heads could support a representation of the \emph{latent contexts} giving rise to the different successor relationships between the tokens in the sequence. With controlled ablation experiments we confirm that these heads support the in-context learning ability of induction heads. Our results are shown for LLMs in the Qwen$2.5$ family of models \citep{yang2024qwen2}, and reproduced for four other models, Gemma$2$-$2$B \citep{team2024gemma}, Llama$3.2$-$3$B \citep{dubey2024llama}, SmolLM$3$-$3$B \citep{bakouch2025smollm3} and Qwen$3$-$0.6$B \citep{yang2025qwen3} in Appendix \ref{appendix:reproducability}.

\section{Task}\label{section:Task}

\subsection{Synthetic data}

To investigate how LLMs learn hierarchical structure in-context, we design token sequences where repetitions appear at various levels of hierarchy. In the simplest case, let us consider a sequence $C = <a,\  b, \  c>$ which is repeated $N$ times to form $C' = <C_1, ..., C_N>$. Induction heads are perfectly fit to capture sequences like these. Each token has a unique successor token, and is repeated in a completely predictable matter without the need to consider higher-order dependencies. 

Next, let us incorporate higher-order dependencies, e.g. where successor tokens can be predicted only by additionally considering the tokens that immediately precede them. Suppose we have three different token sequences like the one designed above $\alpha = <a, \ b, \ c>$, $\beta=<b,\ c, \ a>$, $\gamma=<c,\ b,\ a>$. We refer to these token sequences as \emph{2nd order chunks}. Now we can construct a new sequence where we randomly transition between our 2nd order chunks $\alpha$, $\beta$ and $\gamma$. Since these consist of the same tokens in different orders, simply attending to any successor token will not be a successful strategy for prediction. For instance, the $c$ token is succeeded by $a$ in the $\beta$ chunk, and by $b$ in the $\gamma$ chunk. To this end, if induction heads are responsible for in-context learning in these cases, they have to learn what successor tokens to attend to. We refer to sequences composed of 2nd order chunks as \emph{2nd order} sequences.

Finally, let us consider repetition at one more level of hierarchy. Like in the previous paragraph, we can treat the building blocks of our 2nd order sequences $\alpha, \ \beta, \ \gamma$ as primitive and compose a more complex vocabulary from them. Consider the set of \emph{3rd order chunks} as follows: $\phi = <\alpha, \ \gamma, \ \beta>, \ \psi = <\beta, \ \alpha, \ \gamma>, \ \Omega = <\gamma, \ \alpha, \ \beta>$. We can compose a new sequence with predictable, repetitive patterns by randomly transitioning between the 3rd order chunks. Here, too, simply attending uniformly to successor tokens is futile: For tokens embedded in the 2nd order chunks $\alpha, \ \beta$ or $\gamma$, the induction head has to attend to successor tokens from the same 2nd order structures. For tokens that transition \emph{between} the 2nd order structures, the induction heads need to attend to successor tokens in the right position within the same 3rd order chunk $\phi, \ \psi$ and $\Omega$. We refer to sequences composed of 3rd order chunks as 3rd order sequences.

Throughout this paper we evaluate models on sequences in these three levels of hierarchy, with a focus on the last two. To create them, we fix a small vocabulary, a randomly sampled subset consisting of $V$ tokens drawn from the LLM's original vocabulary. We then generate $P$ unique permutations of the new vocabulary, resulting in $P$ sequences of length $V$. To create what we call a 2nd order sequence, we repeat each permutations $N$ times and shuffle their order (while keeping the 2nd order chunks intact), giving us a total sequence length of $N \times P \times V$. 
Finally, for the 3rd order chunks we construct $P'$ unique permutations of the 2nd order chunks. We then compose a new sequence by randomly shuffling $N$ repetitions of the 3rd order chunks, giving us a total sequence length of $N \times P' \times P \times V$. Example prompts for both 2nd and 3rd order sequences are shown in Appendix \ref{appendix:prompt}.


\section{Induction heads learn to attend in-context}

\begin{figure}[t]
    \begin{center}
    \includegraphics[width=\textwidth]{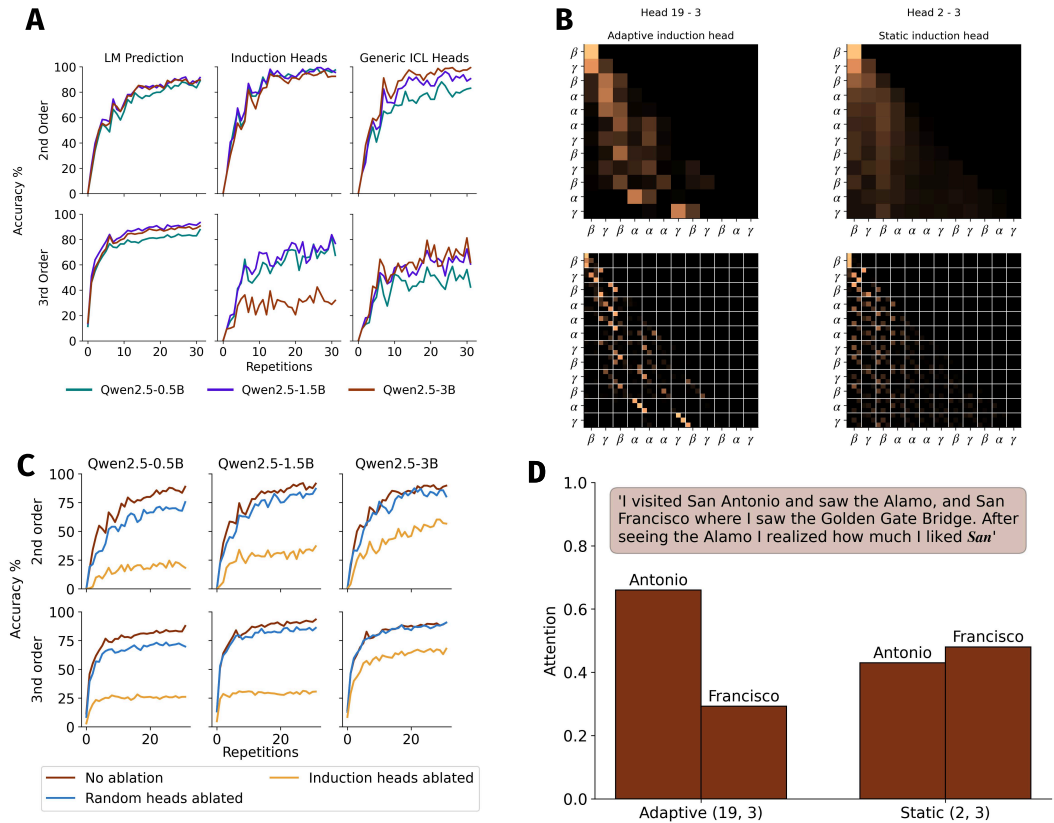}
    \end{center}
    \caption{Some, but not all, induction heads learn to attend to successor tokens in higher-order contexts. (\textbf{A}) Accuracy on the 2nd-order (top row) and the more complex 3rd-order (bottom row) in-context learning tasks. All models can predict the transitions that require an understanding of contexts. There are both induction heads and non-induction heads that learn to attend to the correct context. The learning curves of the heads are the averages of the best performing $5$ heads. (\textbf{B}) Example attention maps showcasing two distinct strategies. The adaptive head (left) successfully learns to attend to the correct successor tokens from previous contextual chunks. In contrast, the static head (right) fails, defaulting to a generic, unfocused attention pattern. (\textbf{C}) Ablating the induction heads severely reduced predictive accuracy and in-context learning for all LLMs across the 2nd and 3rd order sequences, indicating that these are key for in-context learning of repetitive patterns with hierarchical structure. (\textbf{D}) Adaptive induction heads identify the correct successor token in natural language as well. Static induction heads that appear earlier do not and attend more uniformly to the two possible successor tokens. Accuracies in \textbf{A} and \textbf{C} are averaged across 32 different sequences.}
    \label{fig:learning}
\end{figure}

We evaluated three LLMs from the Qwen2.5 family \citep{yang2024qwen2}, with $0.5$, $1.5$ and $3$ billion parameters, on the synthetically generated sequences from the three levels of hierarchy described in Section \ref{section:Task}. Notably, since the sequences were composed of random tokens from the vocabulary and were generated using the procedure above, they never resembled natural language text. Despite this, all models learned to predict the sequences accurately through in-context learning (see Fig. \ref{fig:learning} \textbf{A}). This suggests that there are dedicated circuits for discovering and predicting structured data patterns, even if the data do not resemble natural language.

Next, we assessed if the models had induction heads that learned to attend to the \emph{correct} successor tokens in-context. To do this we first determined which of a model's attention heads were induction heads. We did this using a conventional score matching procedure, where each attention head's attention matrix was matched with an ideal induction head mask for a sequence with random tokens repeated twice \citep{olsson2022context}.

This analysis left us with a pool of induction heads per model. We then assessed whether these heads made tokens attend to their successor tokens from the correct contexts. For 2nd order sequences, we calculated how often a token $a_t$ attended to a successor of $a_{t'<t}$ where $a_{t'<t}$ was embedded in the same type of 2nd order chunk. For 3rd order sequences, we designed a stricter criterion for learning. We calculated whether tokens that marked predictable transitions between 2nd order chunks attended to successor tokens from previous instances of the same 3rd order chunk.

Remarkably, in all models we evaluated we discovered several induction heads that learned to attend to the correct successor tokens in-context, based on the above criterion for correctness. This shows that more sophisticated induction heads exist in LLMs. Induction heads that learned in-context were located in later layers, whereas induction heads that did not show signs of learning were embedded in earlier layers. We showcase example attention maps of both adaptive, or learning, induction heads as well as their static counterparts in Fig. \ref{fig:learning}\textbf{B}.

\subsection{Induction head learning explains in-context learning in our task}

Next we assessed whether it was the activity of the induction heads that indeed was responsible for the LLMs learning to predict our synthetic token sequences in-context. To test this we ablated all induction heads by setting the output of their self attention mechanism to a vector of zeros. As a control condition we repeated this experiment, but ablated an equal number of randomly sampled non-induction heads. We confirmed that the induction heads were indeed responsible for the in-context learning in the LLMs. When they were ablated, the $0.5$B and $1.5$B parameter models predicted tokens at chance level. The  $3$B parameter model was still better than chance after the ablation, but its performance was severely reduced. The control ablation on the other hand produced small or negligible reductions in accuracy (see Fig. \ref{fig:learning}\textbf{C}). 

\subsection{Natural language example}
\label{subsec:nl}

Finally, to verify that induction heads with an adaptive attention mechanism are involved for predicting natural language, we further evaluate the Qwen$2.5$-$1.5$B on a simple sentence construct. Again, a pervasive problem in natural language is that certain tokens like the articles \texttt{the} or \texttt{a}, or common prefixes like \texttt{San} (as in San Francisco) or \texttt{New} (as in New York), tend to have many possible successor tokens even within a single text. A representative sentence could be something like the following:

\begin{tcolorbox}[rounded corners, colback=white,colframe=black, width=\textwidth, title=\textbf{Prompt} \label{prompt:prompt}]
 
I visited San Antonio and saw the Alamo, and San Francisco where I saw the Golden Gate bridge. After seeing the Alamo I realized how much I liked San \texttt{[PREDICTION]}
\end{tcolorbox}

If induction heads are to aid in the prediction process here, the LLMs need to use contextual information (either provided in the prompt, or memorized through pretraining) to direct the \texttt{SAN} token to attend to one of the \texttt{ANTONIO} tokens. We inspected the attention scores for the tokens in this sequence for two induction heads: Head 19-3\footnote{We refer to attention heads using the following scheme: \emph{layer index} - \emph{head index}. For both layer and head indices we use 0-indexing.}, which showed strong learning scores in the synthetic data prediction task, and Head 2-3, which showed no learning. Consistent with our previous results, we see that the adaptive induction head attended more to the \texttt{ANTONIO} token than the \texttt{FRANCISCO} token, whereas Head 2-3 attended more or less equally to the two successor tokens (see Fig. \ref{fig:learning}). We verify the robustness of these results by presenting aggregate evidence over $32$ different prompts in Appendix \ref{appendix:natural_language}.

\section{Learning the building blocks of the hierarchy}
\label{sec:learning_hier}

So far we have presented evidence that LLMs learn to predict structured, repetitive patterns with higher-order dependencies in-context using induction heads that learn what to attend to in-context. But if we explain in-context learning through yet another in-context learning mechanism, we may wonder if we have a satisfactory explanatory account of the phenomenon. In fact, we may wonder how \emph{induction heads themselves learn in-context}?

We propose a mechanism for explaining how induction heads learn what to attend to in our task. Consider the 2nd order sequences we evaluate the models on. The transition relationship between the tokens can be characterized in terms of the transition structure of the $P$ 2nd order chunks $\alpha, \ \beta$, etc. For the 3rd order sequences, the transition relationships are determined both by the 2nd order chunks and the 3rd order chunks. If the model learns to represent that a token belongs to one of these $P$ latent contexts, it can use this to produce keys and queries that allow induction heads to attend to appropriate successor tokens.

To assess whether the models become aware of the underlying, latent contexts that determine the token-transition relationships, we trained linear probes to decode from the models' token representation whether the tokens that belonged to a particular latent context (say, $\alpha$ for 2nd order sequences) were identical to the tokens of the previous latent context.

We trained probes to decode this binary variable from token representations associated with each attention head $\mathbf{z}$. Specifically, each attention head produces a representation for each token $\mathbf{z}_i$, which is the sum of all $J$ tokens' value vectors $\mathbf{v}_j$ multiplied with how much token $i$ attends to token $j$ in that particular head $a_{i,j}$:
\begin{equation}\label{eq:attn_rep}
    \textbf{z}_{i} = \sum_{j=0}^{J} a_{i, j} \mathbf{v}_j
\end{equation}

After training probes to decode these latent context identities from $\mathbf{z}$ (averaged within each context), we evaluated the probes on a left-out test set. High decoding accuracy meant that these representations contained information about whether the previous $n$-order chunk was the same as the current $n$-order chunk.

\begin{figure*}[h]
    \begin{center}
    \includegraphics[width=\textwidth]{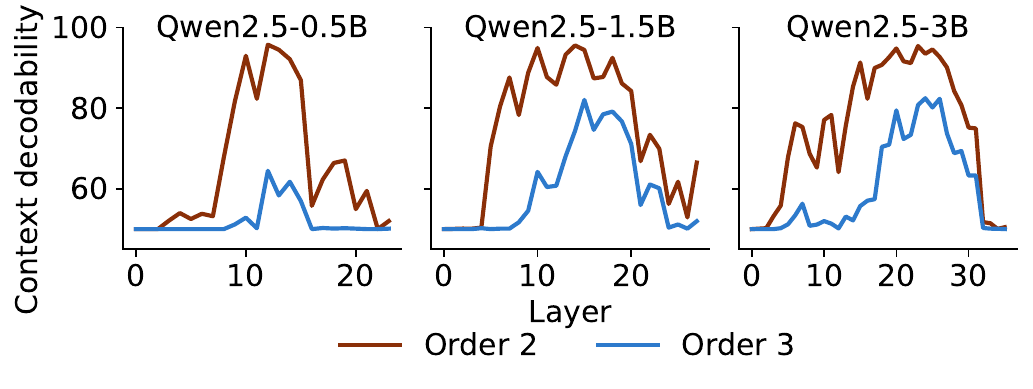}
    \end{center}
    \caption{We trained linear probes to decode from each attention head's representations (see \eqref{eq:attn_rep}) whether the \emph{previous} 2nd or 3rd order chunk was of the same type as the one the token was embedded in. Lines represent the max test decodability over all the heads in a particular layer. Decodability for 2nd order chunk identities was high for all models. For 3rd order chunk identities we see better decodability with model size.}
    \label{fig:decodability}

\end{figure*}

Notably, our analysis revealed that many heads produced representations encoding these 2nd- and 3rd-order chunk identities (see Fig. \ref{fig:decodability}). We name such attention heads \emph{context matching heads}.
In context matching heads, we saw that 2nd-order chunk decodability was above 90\% for several attention heads. 3rd-order chunk decodability was lower, but still substantially higher than chance, and always emerged in context matching heads located in later layers than the heads with high 2nd-order decodability. This makes sense as the models had to build up representations of the 2nd-order hierarchy before being able to build representations of the 3rd-order hierarchy.

\subsection{Building blocks are learned by context matching heads}
\label{subsec:building}

\begin{figure*}[t]
    \begin{center}
    \includegraphics[width=\textwidth]{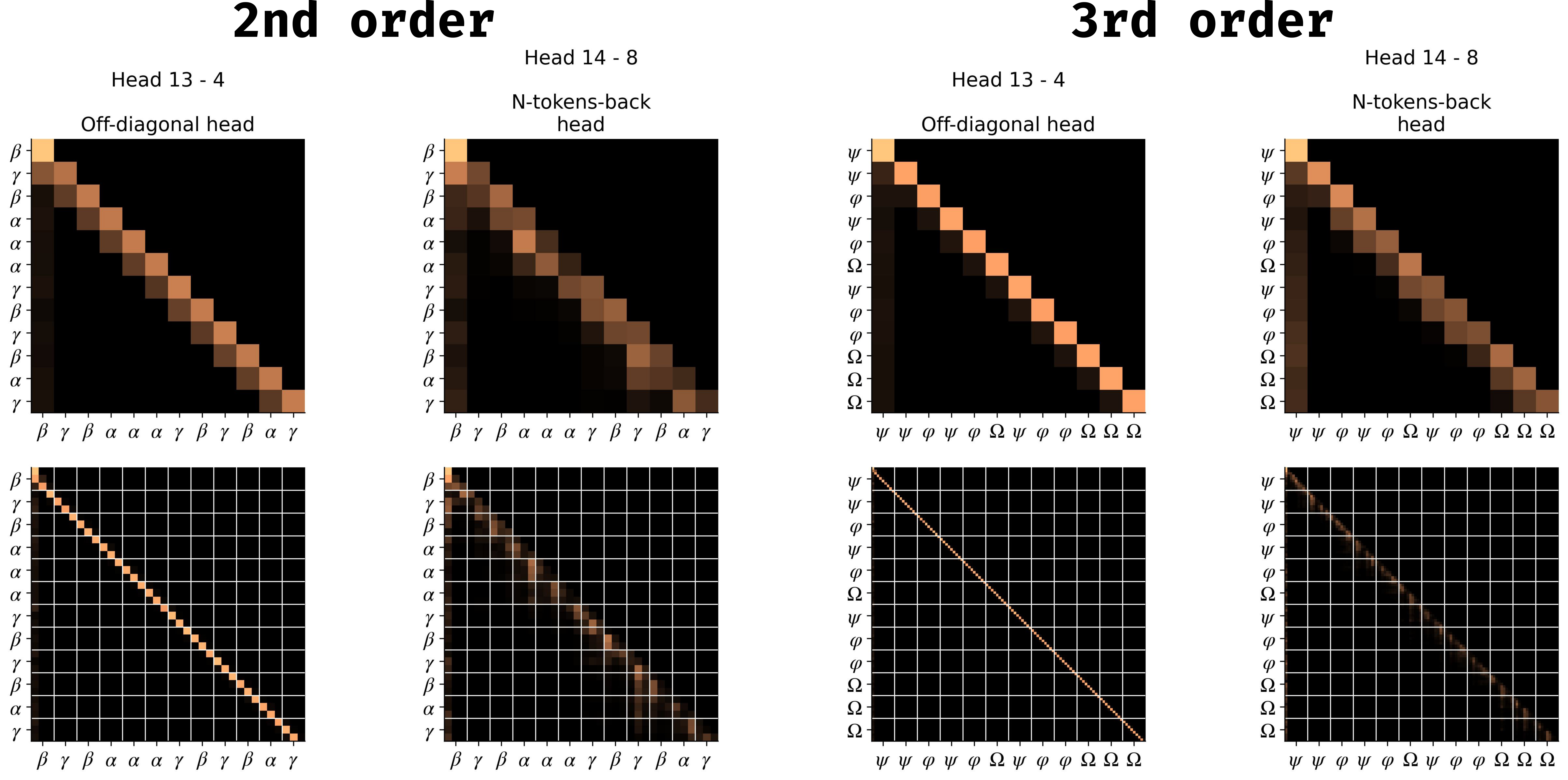}
    \end{center}
    \caption{A showcase of two attention heads with high latent context identity decodability for Qwen$2.5$-$1.5$B. Head 13-4 invariably makes each token attend to its direct predecessor token. By propagating information about the previous token forward, induction heads can easily make tokens attend to their successors in the past. Head 14-8 on the other hand propagates information from longer chains of tokens forward, potentially allowing induction heads to match $n$-grams before making tokens attend to particular successor token from appropriate contexts.}
    \label{fig:one_back_heads}
    
\end{figure*}

If there are attention heads that encode 2nd and 3rd order chunk identities, what do their corresponding attention maps look like? In Fig. \ref{fig:one_back_heads} we visualize two attention heads from Qwen$2.5$-$1.5$B where the 2nd order chunk identity could be decoded with a test accuracy of $> 90\%$. We constructed shorter 2nd and 3rd order sequences using the procedure described in Section \ref{section:Task} and inspected the attention heatmaps. One of the heads (layer $13$, head $4$) made each token attend to its predecessor token. Such heads have previously been found to pair with induction heads, and have been theorized to \emph{copy} over representations of the previous token to its successor to enable the induction head mechanism. However, our analyses suggest that these heads could also be implicated in building up $n$-gram statistics by iteratively routing information from past tokens, making up the current token's latent context. The fact that one can decode the 2nd and 3rd order chunk identities from them suggest that they could enjoy a more general functionality - routing higher-order contextual cues successively from potentially distant tokens to help disambiguate what successor tokens induction heads should attend to. Furthermore, we also discover a variant of such heads wherein tokens not only attend to their direct predecessor, but also to the $N$ previous tokens that precede them (Fig. \ref{fig:one_back_heads}, Head 14-8).

\begin{figure*}[t]
    \begin{center}
    \includegraphics[width=\textwidth]{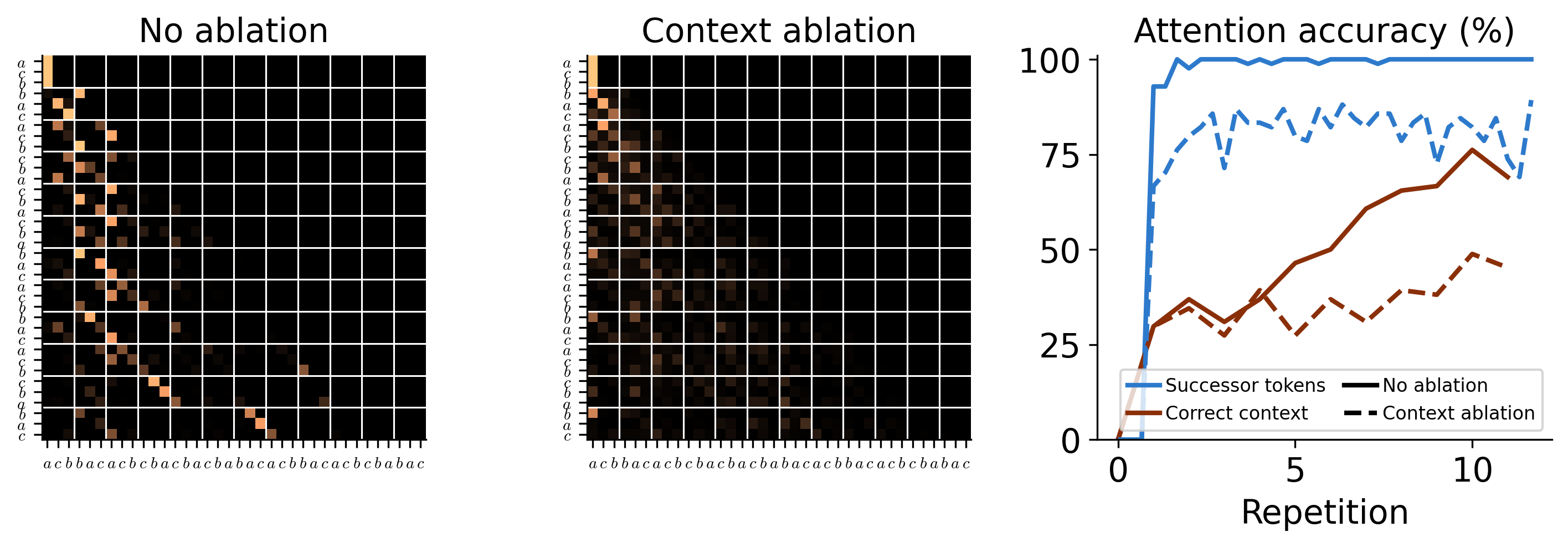}
    \end{center}
    \caption{When ablating a single context matching head, we observed a substantial drop in the subsequent induction head's ability to attend to successor tokens in the correct context. However, the induction head's ability to attend to \emph{successor tokens} in general remained mostly intact. The two heatmaps show attention patterns for head 14-3 for an example sequence. Lines represent the mean from $84$ randomly generated 2nd order sequences. Chance level was $33\%$.}
    \label{fig:one_back_heads_abl}
    
\end{figure*}

We hypothesized that attention heads like these were responsible for creating representations of the latent contexts, enabling induction heads to attend to successor tokens from the appropriate contexts. To test this hypothesis, we designed another ablation experiment where we observed how individual induction heads behaved after ablating a context matching head that directly preceded them. Specifically, we ablated head 13-4 in Qwen$2.5$-$1.5$B, which showed an off-diagonal, look-one-token-back pattern and had a latent context decodability accuracy of $>90\%$. We then observed the behavior of induction head 14-3, located in the subsequent layer. If this head was only copying over the previous token information, ablating it should only affect the subsequent induction head's ability to attend to successor tokens. However, upon ablating this head, we observed that the subsequent induction head still predominantly attended to successor tokens, but almost completely lost its ability to attend to successor tokens from the \emph{right context}. This suggests that the one-token-back attention mechanism is responsible for integrating higher-order contextual cues. This makes sense, as chains of one-token-back attention heads can successively route more distal context forward, similar to a sliding-window attention mechanism \citep{beltagy2020longformer}. These results are shown in Fig. \ref{fig:one_back_heads_abl}.

\subsection{In-context learning suffers from ablating context matching heads}

To assess more broadly whether context matching heads were responsible for the in-context learning we observed in the induction heads, we ran controlled ablation experiments. In these ablation experiments we zeroed out the representations of the context matching heads (any attention head whose latent context decoding score was higher than $85\%$), and observed how these interventions affected the prediction accuracy of the LLM as well as the attention accuracy of its induction heads. To obtain a controlled comparison, we conducted separate ablation experiments where we zeroed out activations of an equal number of randomly sampled attention heads (excluding the context matching heads whose latent context decodability was higher than $55\%$). This allowed us to directly compare the effect of ablating the context matching heads vs a random population of heads with different functionalities. We report learning scores averaged across 32 samples. In the control condition, we randomly picked a set of attention heads to ablate for each of the 32 samples.

\begin{figure*}[t]
    \begin{center}
    \includegraphics[width=\textwidth]{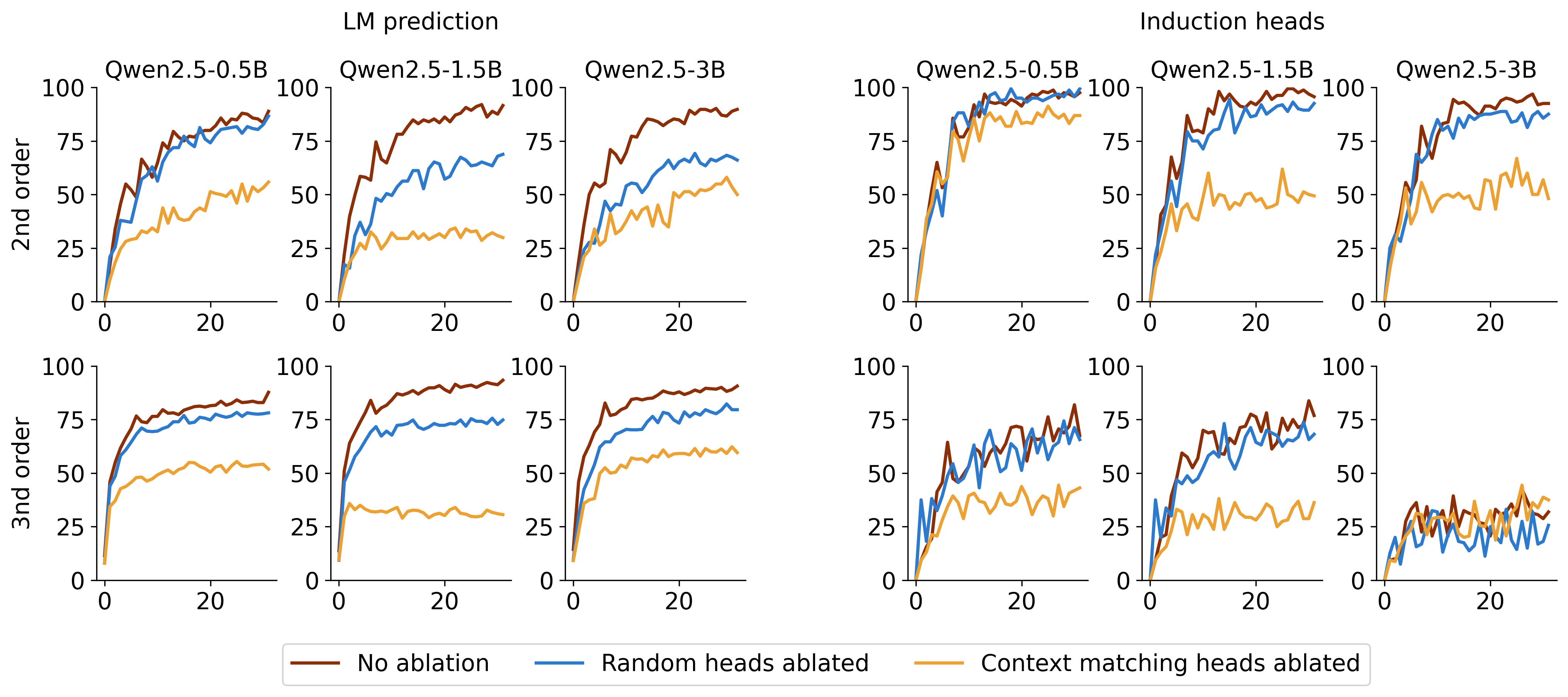}
    \end{center}
    \caption{(\textbf{Left}) LLM predictive accuracy in-context suffers from ablating context matching heads, but much less so from ablating random pools of non-context matching heads of equal size. This trend is observed across all models. (\textbf{Right}) Consistent with our previous results, we also see a notable reduction in the accuracy with which induction heads attend to successor tokens from the correct contexts. Lines represent mean from $32$ independently generated sequences.}
    \label{fig:context_ablation}
\end{figure*}

We observed a consistent and sharp reduction in prediction accuracy when we ablated the context matching heads (see Fig. \ref{fig:context_ablation}, left). In comparison, ablating the same number of randomly picked attention heads that did not encode the latent context produced much smaller adverse effects on prediction accuracy. For Qwen$2.5$-$0.5$B, the adverse effects were almost negligible when we ablated non-context matching heads.

Next we analyzed how these ablations affected the behavior of the induction heads. Here, too, we observed consistent drops in the in-context learning ability of induction heads across both 2nd and 3rd order sequences, suggesting that induction heads learn with the help of context matching heads, making tokens peer back at previous tokens to infer the set of latent token-to-token transition relationships (see Fig. \ref{fig:context_ablation}, right).

\section{Related work}

\textbf{Analyses of in-context learning.} In-context learning, one of the most pervasive features of modern language models, has been analyzed from various theoretical and empirical points of view. \citet{brown2020language} showed that Transformer language models were generally better at predicting tokens with more examples provided as context across a variety of tasks. Subsequent work focused on modeling the phenomenon of in-context learning in Language models as gradient descent \citep{von2023transformers}, Bayesian inference \citep{xie2021explanation}, latent variable inference \citep{hendel2023context}, as well as in virtue of specific circuits like induction heads \citep{elhage2021mathematical, singh2023transient} and training distribution \citep{chan2022data}. Recent work has also focused on analyzing in-context learning empirically using tools like Sparse Auto-Encoders and probing \citep{demircan2025sparse, park2024iclr, park2024competition, akyurek2022learning}.

\textbf{Induction and compression.} Being able to induce repetitive structure is one of the basic building blocks of compression \citep{bartik2015lz4, sayood2017introduction, deletang2023language, saanum2023reinforcement, solomonoff1964formal}. As we have shown, this ability is afforded in part by induction heads and context matching heads. Notably, these induction heads show a remarkable invariance to the particular token sequences they are presented with, picking up on repetitive structure even if the sequences are composed of arbitrary tokens that never co-occur in natural language. Invariances like these are crucial for generalization and for the LLMs ability to serve as general-purpose compressors \citep{olshausen1993neurobiological, saanum2024simplifying, quessard2020learning}.

\textbf{Zoo of attention heads for in-context learning.} Various types of attention heads have been identified to be important for in-context learning. \citet{elhage2021mathematical} and \citet{olsson2022context} identified induction heads that operate at the token level, with some implications of carrying more abstract functions. Later research \citep{yin2025attentionheadsmatterincontext} has argued that in-context learning is also driven by function vector heads \citep{todd2023function}, rather than purely by induction heads. Several other types of head have been identified that support various forms of in-context learning, some of which include concept induction heads \citep{feucht2025dualroutemodelinduction}, semantic induction heads \citep{ren-etal-2024-identifying}, n-gram heads \citep{akyürek2024incontextlanguagelearningarchitectures} and symbolic induction heads \citep{yang2025emergentsymbolicmechanismssupport}. Our work not only discovers a complementary type of attention head, but also highlights the mechanism through which these heads learn, and how this relates to the original induction head circuit (see section ~\ref{subsec:building}).

\textbf{Learning structured sequences.} Lastly, several studies have shown that when LLMs are given sequences generated from latent structures, their internal representations reflect the latent structure \citep{demircan2025sparse, park2024iclr, shai2024transformers}. We build on this line of work by showing that LLMs can learn higher-order transition structures and that they represent key information regarding these structures, such as whether a given higher-order structure matches the preceding one.

\section{Conclusion}

In this paper we have made two important contributions to our understanding of in-context learning in LLMs. 1) We have shown that LLMs can learn repetitive structures with hierarchical dependencies using induction heads that learn what to attend to in-context. 2) We have presented evidence that these induction heads learn through an accompanying circuit of attention heads that serve to discover the latent contexts that give rise to the different token-to-token transition relationships in the input prompt. We observed that these heads make tokens attend to either directly preceding tokens, or longer chains of preceding tokens, routing information allowing subsequent induction heads to query into the successor tokens that have similar chains of preceding tokens. Overall, our results suggest that induction heads can offer a unifying account of how LLMs learn to predict patterns introduced in-context. We have shown that this also holds for prevalent natural language cases, where induction heads can learn to attend to the correct successors of tokens like \texttt{the}, which usually precede conventionally precede multiple different nouns. 

\textbf{Limitations:}
While we have argued that induction heads, with the supporting context matching heads, can give a unifying account of LLMs' ability to induce repetitive structures with hierarchical dependencies, there are other types of in-context learning that we have not explored. For instance, problems from the Abstraction and Reasoning Corpus \citep{chollet2019measure} require LLMs to learn abstract relationships between tokens in few examples. It is unclear if our proposed circuit, relying on context matching, is powerful enough to induce abstract relationships like these. Secondly, all learning analyzed here happened non-verbally, without explicit verbal reasoning. Encouraging in-context learning through reasoning and deliberation may be a second mechanism by which an LLM can change how induction heads allocate attention. Our study paves the way for future work to study induction head behavior in these settings.

\section*{Acknowledgments}

TS and SJG were funded by the Department of Defense MURI program under ARO grant W911NF-23-1-0277. SJG was funded by the Kempner Institute for the Study of Natural and Artificial Intelligence, and a Polymath Award from Schmidt Sciences. ES was funded by the Volkswagen Foundation, the Max Planck Society, the German Federal Ministry of Education and Research (BMBF), and an ERC Starting Grant. TS and CD were funded by the Institute for Human-Centered AI at the Helmholtz Center for Computational
Health. TS was partly funded by Tubingen AI Center, FKZ: 01IS18039A, and funded by the Deutsche Forschungsgemeinschaft (DFG, German Research Foundation) under Germany’s Excellence Strategy–EXC2064/1–390727645.15/18.

\bibliography{iclr2026_conference}
\bibliographystyle{iclr2026_conference}

\newpage

\appendix

\section{Appendix: Implementation details}
\subsection{Synthetic data generation}\label{appendix:task_parameters}

\begin{figure}[h]
    \begin{center}
    \includegraphics[width=\textwidth]{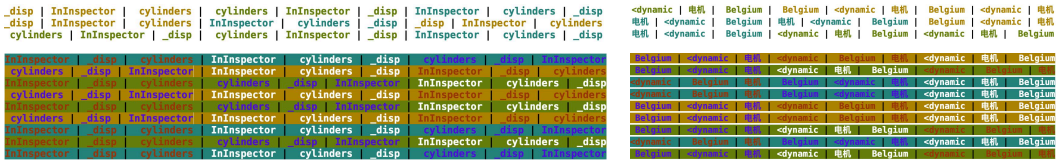}
    \end{center}
    \caption{Example prompts. On the upper row are two second order sequences with three unique tokens, three unique chunks, and three repetitions. Tokens are separated by a vertical bar. The colors of the tokens indicate to which unique chunk they belong to. On the lower row are third order sequences, which additionally include three unique higher order chunks. Here we use different background colors to denote different unique higher order chunks. The sequences are split into new lines in the figure only for illustrative purposes.}
    \label{fig:example_prompt}
\end{figure}

The parameters that were used to generate the synthetic data are shown in Table \ref{table:parameters}. 
The sequence parameters were shared across learning, ablations, and decoding experiments. The $P'$ parameter was only used to generate the $3$rd order sequences. In the 3rd order sequences, we the length of the 2nd order chunks ($V$) was halved to avoid prohibitively long sequences. We used a larger batch size in order to train the linear probes for the decoding analysis. For the qualitative demonstrations (i.e., Fig. \ref{fig:overview}, \ref{fig:learning}B, \ref{fig:one_back_heads}, \ref{fig:one_back_heads_abl}, \ref{fig:example_prompt}) we used shorter sequences. Example prompts for both 2nd and 3rd order sequences are shown in Fig. \ref{fig:example_prompt}.

\begin{table}[htbp]
    \centering
    \begin{tabular}{llllll}
    \toprule
    Experiment & Batch Size & $N$ & $P$ & $P'$ & $V$ \\
    \midrule
    Learning \& Ablation & 32 & 8 & 4 & 4 & 8 (4 for 3rd order sequences) \\
    \addlinespace
    Decoding & 64 & 8 & 4 & 4 & 8 (4 for 3rd order sequences)\\
    \bottomrule
    \end{tabular}
    \caption{The parameters used to generate the synthetic data.}
    \label{table:parameters}
\end{table}

\subsection{Decoding analysis} 
To assess whether the models represented the latent generative contexts, we trained probes to decode from the outputs of each attention head whether a chunk of tokens were generated by the same latent context as the previous chunk of tokens. To obtain the input variables to our decoder we therefore averaged token representations within a context.

the context of the current token was the same as the previous context. Next, we optimized an $L2$-regularized logistic regression model using the \texttt{scikit-learn} library \citep{pedregosa2011scikit}. We used $75\%$ of the data to train the classifier and the remaining $25\%$ to test it. Due to the structure of the sequences, the previous context and the current context were more likely to be different, creating a class imbalance. Therefore, the accuracy scores reported in Fig. \ref{fig:decodability} are balanced accuracy scores.

\subsection{Prompt visualization}\label{appendix:prompt}

\subsection{Natural Language}\label{appendix:natural_language}

We tested whether adaptive induction heads are successful in resolving ambiguities in natural language. We carried out the same analysis presented in Section \ref{subsec:nl}. We provided Qwen$2.5$-$1.5$B with $16$ unique sentences, where the the ambiguity of which token to predict next can only be solved based on the previous context. With counterbalancing the order of the examples, we had $32$ different test cases, whose results are shown in Fig. \ref{fig:nl_appendix}, providing further evidence that adaptive induction heads can infer which token to attend to from the context in ambiguous cases. We provide the prompts that were used in Table \ref{table:prompts}.

\begin{figure}[h]
    \begin{center}
    \includegraphics[width=\textwidth]{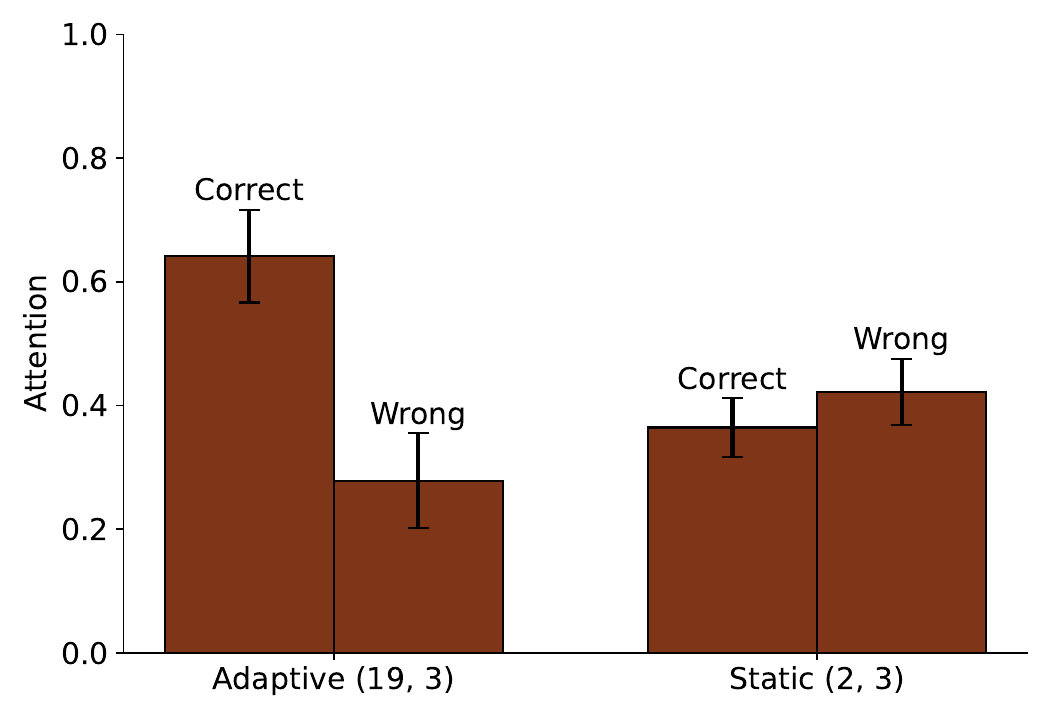}
    \end{center}
    \caption{Adaptive induction heads attend to the correct tokens in ambiguous situations. Static induction heads do not have such a preference.}
    \label{fig:nl_appendix}
\end{figure}

\begin{table}[htbp] 
    \centering 
    \begin{tabularx}{\textwidth}{@{} l X l l @{}}
    \toprule
    \textbf{\#} & \textbf{Prompt} & \textbf{Correct} & \textbf{Wrong} \\
    \midrule
    1 & My cousins are David Chen and David Lee. I needed to speak with the one whose last name is only three letters long, so I called David \dots & Lee & Chen \\
    \addlinespace
    2 & Both the Statute of Liberty in New York and the French Quarter in New Orleans are famous tourist attractions. Because I love southern cuisine, I decided to visit New \dots & Orleans & York \\
    \addlinespace
    3 & The gift box contained an Apple Watch and an Apple iPhone. The item designed to be worn on the wrist was the Apple \dots & Watch & iPhone \\
    \addlinespace
    4 & The report was on the Bank of America and the Bank of Canada. Since the focus was on Canadian financial institutions, I wrote about the Bank of \dots & Canada & America \\
    \addlinespace
    5 & The curriculum covered both World War 1 and World War 2. The exam question was about the earlier of the two conflicts, which was World War \dots & 1 & 2 \\
    \addlinespace
    6 & The hotel room had a king-size bed and a king-size pillow. The large piece of furniture I slept on was the king-size \dots & bed & pillow \\
    \addlinespace
    7 & I like reading both Stephen Hawking and Stephen King. Yesterday, I felt more like reading fiction, so I read Stephen \dots & King & Hawking \\
    \addlinespace
    8 & The lecture contrasted the composers John Lennon and John Williams. I don't like the Beatles, so I focused more on John \dots & Williams & Lennon \\
    \addlinespace
    9 & I visited San Antonio and saw the Alamo, and San Francisco where I saw the Golden Gate Bridge. After seeing the Alamo, I realized how much I liked San \dots & Antonio & Francisco \\
    \addlinespace
    10 & They had oat bar and oat milk for breakfast. I don't like drinking anything in the mornings, so I took the oat \dots & bar & milk \\
    \addlinespace
    11 & My granddad likes flowers and my grandmum likes chocolate. Therefore, I will gift these flower to my grand\dots & dad & mum \\
    \addlinespace
    12 & He asked whether I like best of 5 or best of 3 matches more. I get tired quickly, so I said best of \dots & 3 & 5 \\
    \addlinespace
    13 & Between the two, planet Fulty has less sunlight than planet Julty. Because I like the sun, I like going to planet \dots & Julty & Fulty \\
    \addlinespace
    14 & She could either practice her backhand or backspin. Since she already practices her backhand yesterday, today she worked on her back\dots & spin & hand \\
    \addlinespace
    15 & I split my time between Bad Tölz and Bad Homburg. I love Bad Tölz in the winters. Since it is now July, I am in Bad \dots & Homburg & Tölz \\
    \addlinespace
    16 & I like drinking caffè Americano in the morning and caffè mocha in the afternoon. It is now 3 PM, and I would like to drink caffè \dots & mocha & Americano \\
    \bottomrule
    \end{tabularx}
    \caption{The natural language prompts that were used to test the induction heads. Corresponding correct and wrong answers are also provided.}
    \label{table:prompts}
\end{table}

\newpage

\section{Appendix: Replication with other Language Models}\label{appendix:reproducability}
To assess the general validity of the proposed circuit, we investigated if \emph{other} open-source LLM families showed the signs of the same algorithmic implementation of hierarchical in-context learning. To this end, we conducted a subset of our experiments in exactly the same manner on four new LLMs, Gemma$2$-$2$B \citep{team2024gemma}, Llama$3.2$-$3$B \citep{dubey2024llama}, SmolLM$3$-$3$B \citep{bakouch2025smollm3} and Qwen$3$-$0.6$B \citep{yang2025qwen3}. Specifically, we sought to assess 1) if these models also have induction heads that learn in-context. 2) if these models have context matching heads from which one can decode the latent generative contexts in our task. And 3), if these context matching heads were causally involved in the in-context learning ability of the induction heads. To evaluate the last point, we again ablated all context matching heads whose latent context decodability was higher than $85\%$, and observed how this affected model accuracy and induction head accuracy. We  also ablated random heads as a control condition, exactly like in Section \ref{subsec:building}. The results presented for the Qwen$2.5$ models generally reproduced with striking levels of consistency. All models had induction heads that learned in-context, had heads that encoded the latent generative contexts, and that were causally linked to the in-context learning ability of the LLM and the induction heads more specifically. The results are shown individually for each model below.

\begin{figure}[!htb]
    \begin{center}
    \includegraphics[width=\textwidth]{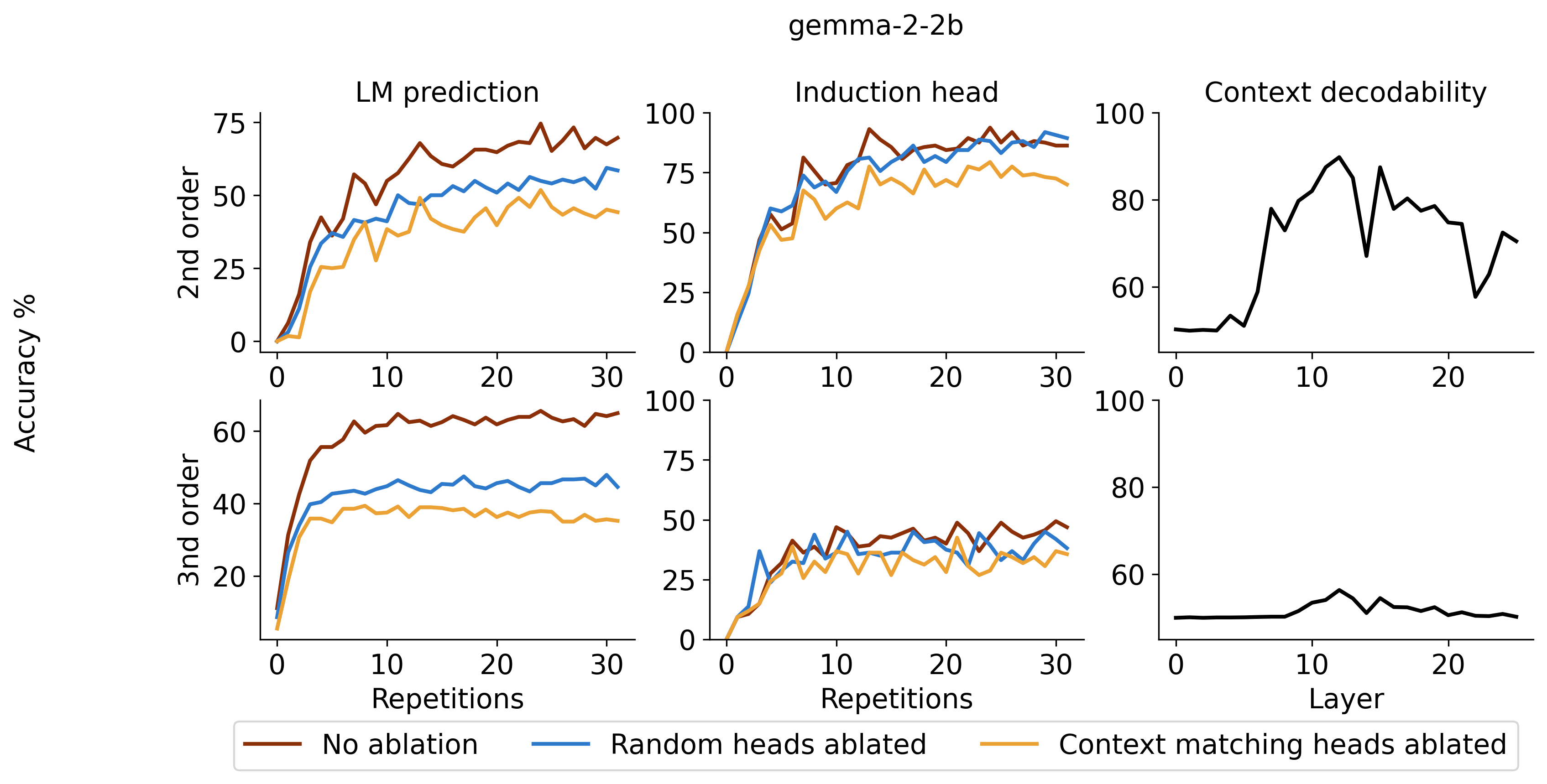}
    \end{center}
    \caption{Experimental evaluation on Gemma$2$-$2$B.}
    \label{fig:nl_appendix}
\end{figure}

\begin{figure}[!htb]
    \begin{center}
    \includegraphics[width=\textwidth]{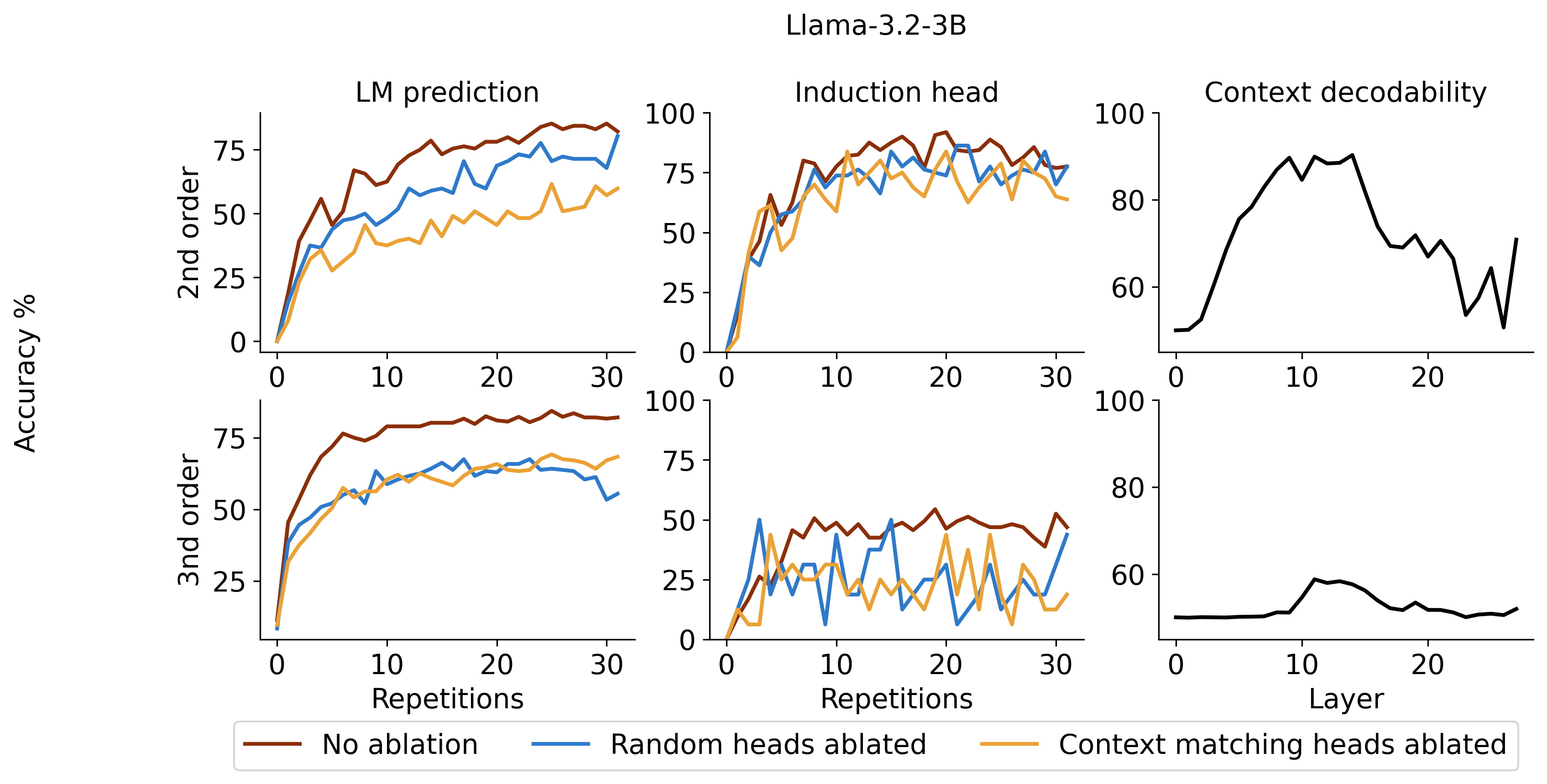}
    \end{center}
    \caption{Experimental evaluation on Llama$3.2$-$3$B.}
    \label{fig:nl_appendix}
\end{figure}

\begin{figure}[!htb]
    \begin{center}
    \includegraphics[width=\textwidth]{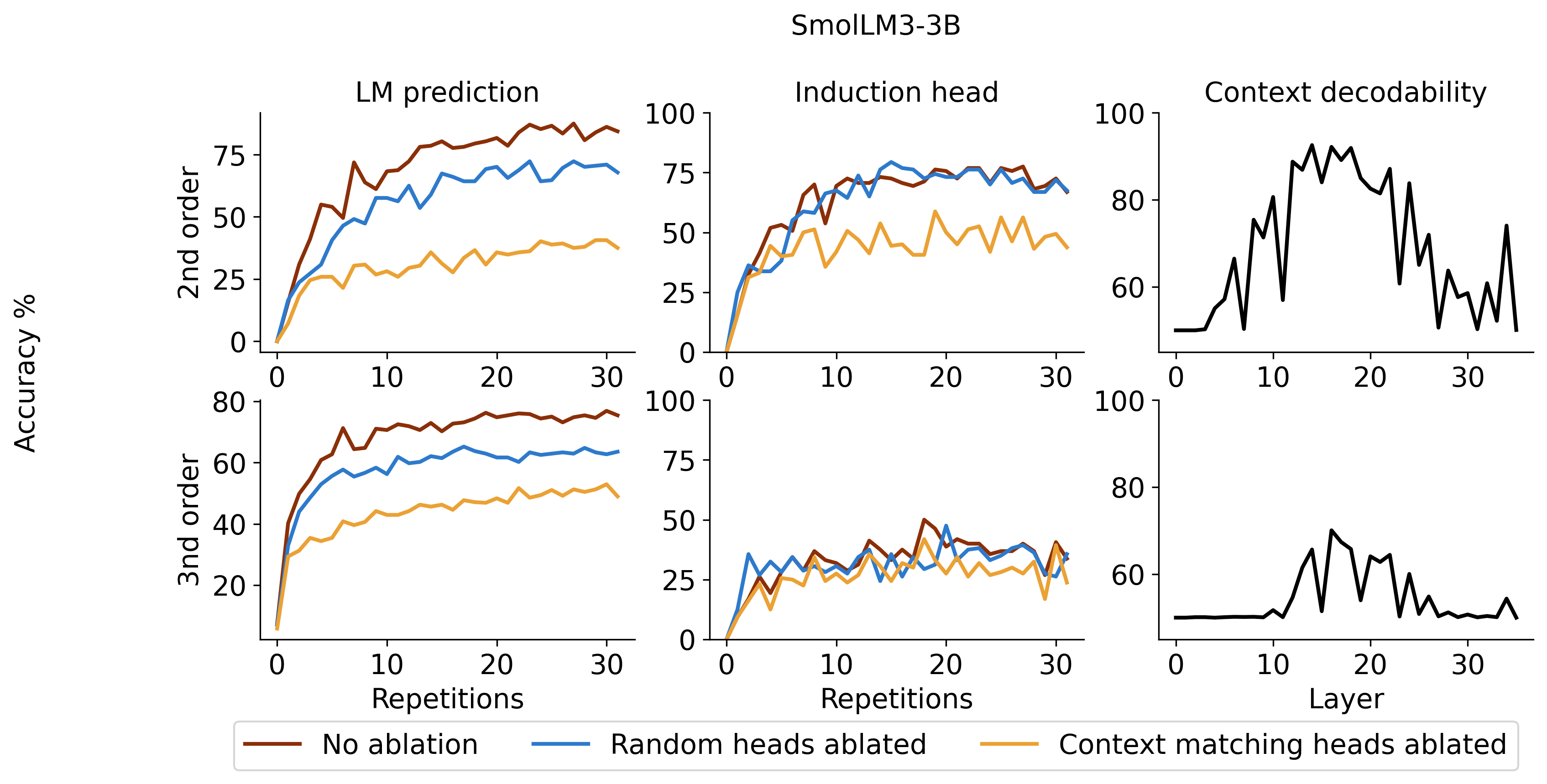}
    \end{center}
    \caption{Experimental evaluation on SmolLM$3$-$3$B.}
    \label{fig:nl_appendix}
\end{figure}

\begin{figure}[t]
    \begin{center}
    \includegraphics[width=\textwidth]{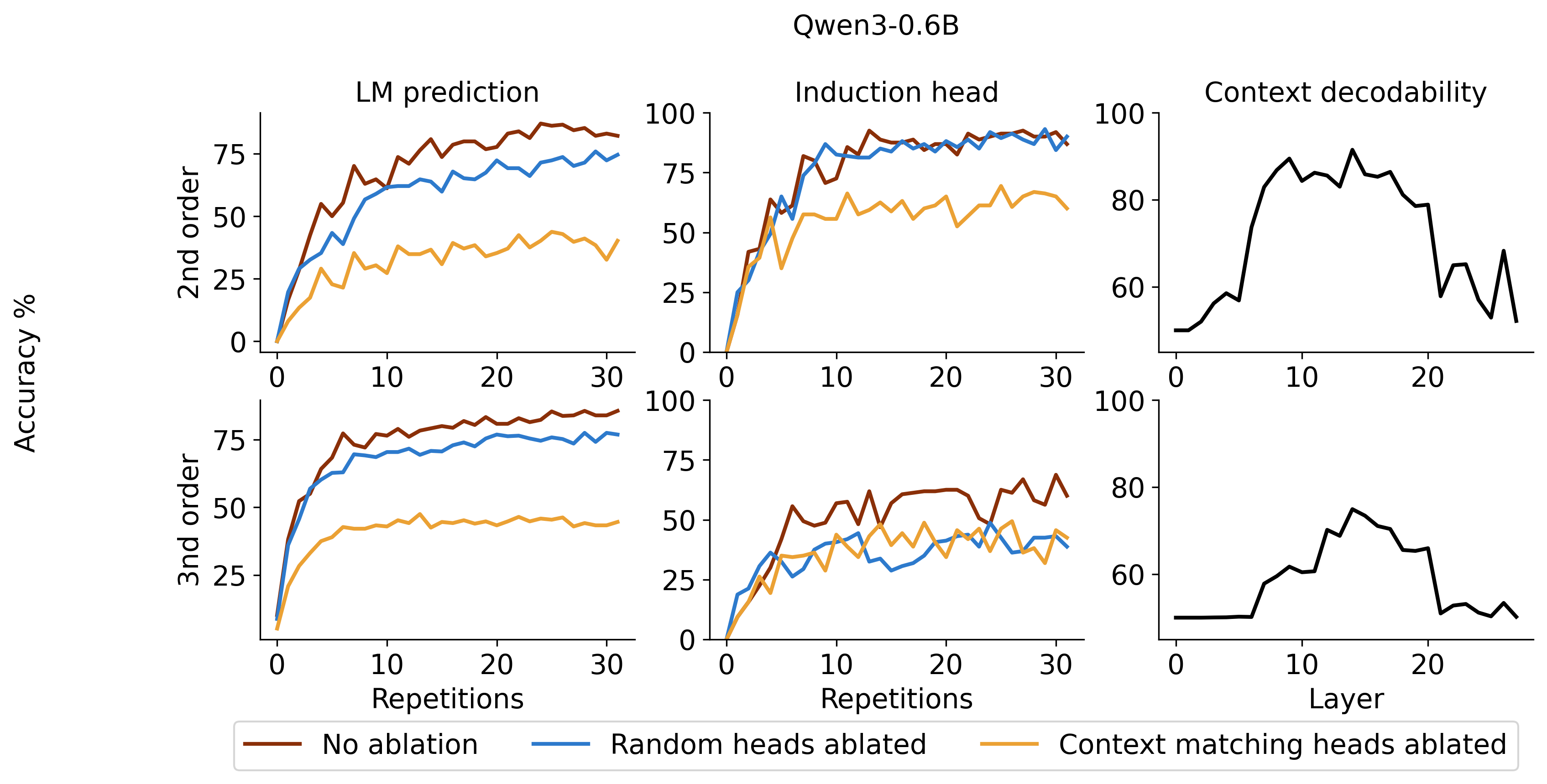}
    \end{center}
    \caption{Experimental evaluation on Qwen$3$-$0.6$B.}
    \label{fig:nl_appendix}
\end{figure}

\end{document}